# Pattern recognition using spiking antiferromagnetic neurons

Hannah Bradley[1] [*], Steven Louis[2], Andrei Slavin[1], and Vasyl Tyberkevych[1]

[1]Department of Physics, Oakland University, Rochester, MI 48309
[2]Department of Electrical Engineering, Oakland University, Rochester, MI 48309
*hbradley@oakland.edu

**Abstract**

Spintronic devices offer a promising avenue for the development of nanoscale, energy-efficient artificial neurons for neuromorphic computing. It has previously been shown that with antiferromagnetic (AFM) oscillators, ultra-fast spiking artificial neurons can be made that mimic many unique features of biological neurons. In this work, we train an artificial neural network of AFM neurons to perform pattern recognition. A simple machine learning algorithm called spike pattern association neuron (SPAN), which relies on the temporal position of neuron spikes, is used during training. In under a microsecond of physical time, the AFM neural network is trained to recognize symbols composed from a grid by producing a spike within a specified time window. We further achieve multi-symbol recognition with the addition of an output layer to suppress undesirable spikes. Through the utilization of AFM neurons and the SPAN algorithm, we create a neural network capable of high-accuracy recognition with overall power consumption on the order of picojoules.

**Introduction**

Despite the increase in the computational capability of typical von Neumann architecture, the human brain still outperforms modern computers at classification tasks with a fraction of power consumption [1], [2]. By mimicking brain-like behaviors through hardware-implemented artificial neural networks, neuromorphic chips perform pattern recognition with reduced power consumption and increased efficiency [3].

A biological neural network is comprised of two critical components: the individual processing units called neurons and the synapses that determine their connections. Current neuromorphic chips use silicon-based transistors to make up both of these components [4]. In spite of the fact that transistor-based neuromorphic computing is an improvement over Von Neumann architecture, a number of drawbacks still exist. Mainly, it requires multiple transistors to create one artificial neuron, thereby requiring a large amount of physical area and increasing power consumption.

In recent years, there has been growing interest in the use of spintronic devices for neuromorphic computing. A comprehensive review of different approaches can be found in Refs [5], [6], [7], [8], [9]. These devices, typically made from magnetic materials, exploit various electron spin patterns. By utilizing spin-dependent effects, such as pure spin current— which is decoupled from electrical current and thus avoids many issues like Ohmic loss—spintronic devices operate with significantly lower power requirements than transistor-based counterparts. Spintronic artificial neurons offer several promising features. Apart from possessing intrinsic non-linear dynamics, these neurons can be fabricated at the nanoscale, with each device serving as a single neuron. As a result, power and area requirements are significantly reduced. The creation of artificial neurons has been shown to be possible with domain wall motion [10], [11], [12], skyrmions [13], [14], [15], spin torque nano oscillators [16], [17], [18], and magnetic tunnel junctions [19], [20].

One of the prospective designs of artificial spintronic neurons is based on antiferromagnetic (AFM) spin-Hall oscillators operating in a subcritical regime [21]. These artificial "AFM neurons" generate voltage spikes that closely resemble the action potentials elicited by biological neurons, with properties that include response latency, bursting, and refraction. The main advantages of artificial AFM neurons are their nano-sized footprint, relatively low power consumption, and ultra-high operational speed, generating spikes with a duration on the order of 5 ps [22]. In light of the high speed and low power consumption of AFM neurons, it is important to consider AFM neurons as a possible candidate for post-silicon neuromorphic computer systems.

Until now, the literature shows no attempt to develop a method to perform machine learning with AFM spiking neurons. While a simple neural network employing AFM neurons in memory loops was presented in Ref. [22], that neural network featured copper bridge synapses that carried spikes from neuron to neuron with constant coupling. Thus, it did not have the ability to demonstrate that more complex neural networks based on AFM neurons can be trained for cognitive tasks like pattern recognition and how efficient these networks are in terms of training, recognition time, and power consumption. To a large extent, this problem depends on the efficient implementation of variable spintronic synapses capable of changing inter-neuron connectivity. This need for variable synapses dramatically increases the complexity of a neural network.

A less demanding approach is based on reservoir computing, which only trains synapses connected to the output layer [23], reducing the number of variable synapses needed and simplifying the overall neural network. Several simple learning algorithms are known for reservoir computing. In particular, a supervised learning algorithm called spike pattern association neuron (SPAN) limits the number of output neurons to one, simplifying the neural network even more [24]. The meaningful recognition tasks are possible in the SPAN algorithm by employing temporal encoding and matching outgoing spikes to a desired time.

In this work, we theoretically investigate the possibility of using AFM neurons combined with the SPAN algorithm to create neural networks that recognize symbols generated from a grid of input neurons [25], [26]. We show that, due to the strongly nonlinear and inertial dynamics of AFM neurons, even a single AFM neuron is capable of successfully recognizing various symbols from a 5 × 5 input grid, which is enough to encode various printed symbols. Our simulations show that the total training time of an AFM SPAN neuron can be below 1 μs, while the power consumption during the training is of the order of 30 pJ. This research provides the first demonstration of the ability of AFM neurons to perform learning tasks, thus making clear the potential for using artificial AFM neurons in machine learning applications.

**Methods**

Antiferromagnets (AFM) have two magnetic sublattices orientated in opposing directions. The direction of AFM magnetic sublattices relative to the crystal lattice can be manipulated using spin currents. Usually, this is achieved in spin Hall geometry, in which a layer of heavy metal covers an AFM element. Here, we consider a spin Hall oscillator created from a bilayer of NiO and Pt. A detailed analysis of this device can be found in Ref. [27]. When a DC electric current flows in the heavy metal layer, it induces a perpendicular spin current that penetrates into the AFM.

The most interesting effect of spin current on the AFM dynamics happens when the spin polarization of the spin current is perpendicular to the easy plane of the AFM. In this case, spin-transfer torque induced by a sufficiently large spin current causes the AFM sublattices to rotate in the easy plane [27]. For AFM materials with bi-axial anisotropy, the rotation of the sublattices is not uniform with time. This results in a spin-pumping effect, where the rotation of the sublattices affects the spin state of the adjacent heavy metal layer. A sequence of short spin-pumping spikes are emitted from the AFM at a frequency that can reach the THz range. When the bias current exceeds a threshold value, the AFM sublattices will continuously rotate in the easy plane, resulting in a train of spin-pumping spikes emitted by the AFM. The threshold current needed to achieve this auto-oscillating regime depends on the easy-plane anisotropy of the AFM material and is of the order of $10^8$ A/cm$^2$ for NiO AFM [27].

If the driving current is below the generation threshold, the AFM oscillator will not have enough energy to overcome the anisotropy, but the equilibrium orientation of the AFM sublattices will be moved towards the hard direction in the easy plane. With an additional impulse of current, the AFM magnetizations will surpass the anisotropy energy barrier and perform a single half-turn in the easy plane, which will cause a single spike of the spin-pumping voltage. Due to the THz properties of AFM materials, the width of this voltage spike is on the order of picoseconds. This response of a sub-threshold AFM spin Hall oscillator is similar to the reaction of a biological neuron to an external stimulus. The AFM neurons and their networks also have other properties that resemble biological neural systems, such as refraction and response latency. A comprehensive study of AFM neurons and single-neuron behavior can be found in Refs [21], [22].

As it was shown in Ref. [27], the dynamics of an AFM neuron can be described by the in-plane angle $\phi$ that the Neel vector of the AFM makes with the easy axis of the AFM. Under rather general assumptions, the angle $\phi$ obeys the second-order dynamical equation,

$$\frac{1}{\omega_{ex}}\ddot{\phi} + \alpha\dot{\phi} + \frac{\omega_e}{2}\sin 2\phi = \sigma I, \tag{1}$$

where $\omega_{ex} = 2\pi f_{ex}$ is the exchange frequency of the AFM, $\alpha$ is the effective Gilbert damping constant, $\omega_e = 2\pi f_e$ is the easy axis anisotropy frequency, $\sigma$ is the spin-torque efficiency defined by Eq. (3) in Ref. [27], $I$ is the driving electric current. Further details about the derivation of Eq. (1) can be found in Refs [21], [27]. Note that the spin-pumping signal produced by the AFM is proportional to the angular velocity of the sublattice rotation $\dot{\phi}$. Namely, the inverse spin Hall voltage produced by the AFM neuron can be found as,

$$V = \beta \dot{\phi}, \qquad (2)$$

where the efficiency $\beta = 0.11 \times 10^{-15}\ V \cdot s$ is defined by Eq. (2) in Ref. [22].

In this work, we study the dynamics of a network of interconnected AFM neurons. Each neuron is described by its own phase $\phi_i$ and obeys an equation similar to Eq. (1) with additional terms describing synaptic connections between the neurons:

$$\frac{1}{\omega_{ex}} \ddot{\phi}_i + \alpha \dot{\phi}_i + \frac{\omega_e}{2} \sin 2\phi_i = \sigma I + \sum_{i \neq k} \kappa_{ik} \dot{\phi}_k. \qquad (3)$$

Here, $i$ and $k$ are indices that represent the $i$-th and $k$-th neurons, and $\kappa_{ik}$ represents a matrix of coupling coefficients. Note that the coupling signal produced by the $k$-th neuron is proportional to $\dot{\phi}_k$, in agreement with Eq. (2).

The coupling coefficients that constitute $\kappa_{ik}$ can behave as the synaptic weights in a machine learning system. To a large extent, the challenge of building a fast and efficient neuromorphic computing system depends on the efficient implementation of variable spintronic synapses capable of changing inter-neuron connectivity. The need for variable synapses dramatically increases the complexity of a neuromorphic neural network. This problem is even more serious for AFM neurons since, to fully employ ultra-fast AFM dynamics in neuromorphic hardware, the reaction times of artificial synapses should be on the timescale of AFM neuron dynamics. As AFM neurons spike with a duration that can be less than 5 ps, it should be noted that traditional CMOS technology would severely limit the capabilities of an AFM neural network. To our knowledge, no variable weight synapses have been developed that are suitable to work in conjunction with AFM neurons. As no CMOS or spintronic hardware is capable of being used as variable synapses for AFM neurons, circuit simulations such as SPICE simulations cannot be done. Therefore, in this paper, which primarily focuses on investigating the dynamics of AFM neurons, we did not assume any particular physical model of a synapse. Instead, the simulated synapses are considered to be "ideal" such that they can be adjusted instantaneously and to any value. The simulations presented in this work are the result of direct numerical calculations of the system of equations described by Eq. (3) for the entire AFM neural network.

Nevertheless, it is important to consider how the latency, or synaptic delay, would impact our model. In a previous work [22], copper bridges with fixed dimensions were used to provide constant weight synaptic coupling or fixed connections $\kappa_{ik}$ between AFM neurons. Copper is capable of carrying spin current from one neuron to the next, allowing the output of one neuron to act as the input for a second neuron. The synaptic delay of copper bridges can be found by solving the diffusion equation for spin accumulation in copper. By using standard diffusion parameters for copper [28] and AFM neuron dimensions found in Ref. [22], the synaptic delay for a copper bridge with a length of 100 nm can be found to be about 1.5 ps. We consider this delay to be short enough to have a negligible impact on our system.

There is a remarkable similarity between the equation describing the AFM neuron and that describing the dynamics of a physical pendulum; therefore, each term in Eq. (3) can be characterized by its mechanical analog. As a result, the coefficient of the first term on the left-hand side of Eq. (3) defines an effective mass, indicating that the AFM neuron possesses an effective inertia due to AFM exchange. This inertia results in a delay or response latency between a neuron receiving an input and the resulting output, an effect not found in conventional artificial neurons. When AFM neurons are linked together, such that the output of one neuron acts as the input of the next, the delay is dependent on the coupling strength $\kappa_{ik}$ between the neurons. The delay caused by inertia decreases as the coupling strength between neurons increases. This is illustrated in Fig. 1, which plots the output voltage spikes, described by Eq. (2), of two AFM neurons connected in a chain. In Fig. 1(a), the two neurons are connected with the coupling coefficient of $\kappa_0 = 0.011$, which was chosen from Ref. [22] where static chains of AFM neurons were studied. This results in the blue neuron firing with a delay of 100 ps. However, when the coupling coefficient is increased to $1.5\kappa_0$, the blue neuron fires with a shortened delay of 50 ps. Thus, the firing time of the neuron can be easily controlled through changes in the synaptic coupling, demonstrating that AFM neurons are well-suited for neuromorphic algorithms that utilize time encoding of neuron spikes.

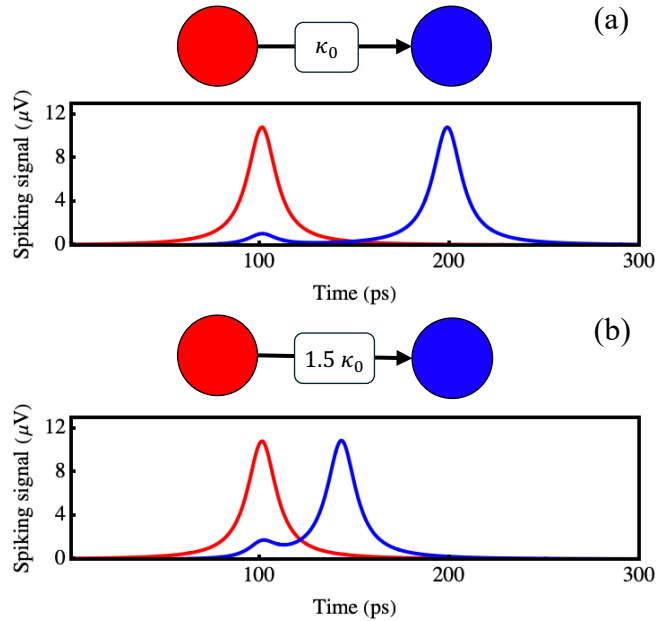

Figure 1. Variable response latency. The simulation results plot the output spiking voltage of the AFM neuron following Eq. (2). (a) Two AFM neurons are coupled with a coupling coefficient of $\kappa_0 = 0.011$. The blue neuron fires with a delay of 100 ps. (b) Two AFM neurons are coupled with a coupling coefficient of $1.5\kappa_0$. The blue neuron fires with a shortened delay of 50 ps.

Several machine learning algorithms utilize this time-encoding approach. A well-known example is Spike-Time-Dependent Plasticity (STDP), an unsupervised learning algorithm that updates weights based on the spike timing of pre- and post-synaptic neurons. Unsupervised learning algorithms like STDP, which use unlabeled training data, are ideal for discovering hidden underlying patterns in large data sets. Alternatively, supervised learning algorithms use labeled data and are ideal for classification tasks such as pattern recognition. This paper examines one such supervised temporal-encoding approach: Spike Pattern Association Neuron (SPAN) [25].

The architecture of an AFM neural network realizing the SPAN algorithm is shown in Fig. 2(a). It consists of one output "SPAN" neuron connected to many neurons of the input layer. In our simulations, the input layer consisted of 25 neurons and encoding input symbols drawn in a 5 × 5 binary grid. We used several shapes of the input symbols shown in Fig. 2(b). A blackened pixel in the input symbol causes a spike in the corresponding input neuron, while a white pixel will have no spike. The SPAN neuron is trained to output a spike at a certain prescribed time if the input symbol matches the pattern to be recognized. To achieve this, synaptic connections between the input layer and the SPAN neuron are adjusted during the training, as explained below.

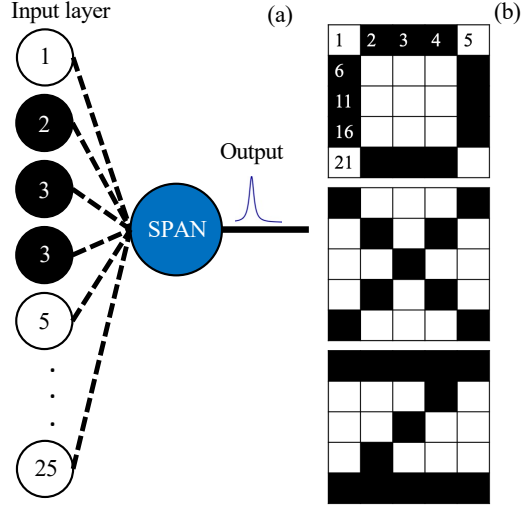

Figure 2. Single AFM SPAN neural network. (a) The architecture of AFM SPAN neural network. The input symbol is encoded in the spiking pattern of the input layer neurons. The synaptic weights between the input layer and the AFM SPAN neuron are adjusted during training. The temporal position of the SPAN's output spike encodes the outcome of pattern recognition. (b) Examples of the different input symbols used in pattern recognition where each pixel cell represents a different input neuron that will fire if the pixel is black.

We used parallel encoding of the input layer; namely, the input symbol triggers the input neurons to fire simultaneously. If the combined weights connected to the SPAN are strong enough, there will be an output spike. The goal of training is to move this output spike to the desired time for a chosen symbol. If the spike is produced earlier (later) than the target time, the weights connected to the SPAN should reduce (increase).

In more detail, the SPAN training algorithm is based on the Widrow-Hoff rule, where the difference between the desired spike time $t_d$ and the actual spike time $t_a$ is used to update the synaptic weights. After some manipulation, shown in Ref. [29], the Widrow-Hoff rule is transformed to describe the change in weights during training:

$$\Delta\kappa = \lambda \left(\frac{e}{2}\right)^2 \left[(t_d - t_i + \tau)e^{-(t_d-t_i)/\tau} - (t_a - t_i + \tau)e^{-(t_a-t_i)/\tau}\right], \tag{4}$$

where $\lambda = 0.05$ is a positive and constant learning rate, $t_i$ is the timing of the input spike, $t_d$ is the desired timing of the output spike, $t_a$ is the actual timing of the output spike, and $\tau = 100$ ps is a time constant corresponding to the width of a spike. Due to the simplicity of the SPAN algorithm, it is only capable of training a neuron to a single symbol. Upon training, a SPAN should output its spike at the target time $t_d$ for the correct symbol and spike away from the target time for any other symbol.

A library of 20 symbols is used to train the neural network. These symbols are all variations of the correct symbol chosen from one of the symbols shown in Fig. 2(b). Variations include symbols with multiple additional or missing pixels. Initially initialized with random synaptic weights $\kappa_{ik}$, the neural network receives each symbol as an input during one training epoch. A symbol variation is associated with a target time corresponding to the image's difference from the correct symbol. Each additional incorrect pixel shifts the target time by 10 ps. For example, the target time for the correct symbol is 100 ps, while a symbol with 2 additional incorrect pixels has a target time of 80 ps. Using this time and the actual timing of the output neuron, the SPAN algorithm determines how the weights should be changed in accordance with Eq. (4). The algorithm is modified to ensure that the weights cannot become negative. Negative weights would necessitate an inversion of spin current in the synapses between neurons, likely requiring CMOS technology, which would significantly increase the complexity of the neural network, negating the advantages of using AFM neurons. Therefore, to maximize the efficiency of the neural network, the weights are kept positive throughout the entire training process. When all images have been processed, the weight changes resulting from each symbol are averaged, the neural network is updated by changing the values of $\kappa_{ik}$ in Eq. (3), and the next epoch begins.

Figure 3 shows the output spikes of a SPAN neural network after training. When the correct symbol is supplied as input, the SPAN spikes within a 10 ps time window of the target time; this implies that the neural network has recognized chosen symbol. Any other symbol should cause a spike outside the target time window, indicating that a different symbol was used as input.

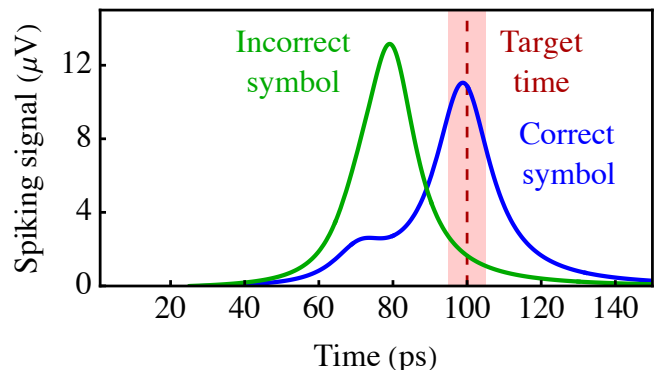

Figure 3. Simulation result of output signals generated by a trained AFM SPAN network. The red dashed line shows the target time for symbol recognition, with a 10 ps time window (red shading) encompassing the target time. The blue (green) line shows the simulated output spike of the AFM SPAN for the correct (incorrect) input symbol.

**Results and Discussion**

      Figure 4(a) shows the error between the actual and desired spike time for the correct symbol throughout training, and Fig. 4(b) shows the corresponding changes in each synaptic weight. In this case, the neural network is being trained to the "O" symbol. After an aggressive start, the change in weights is subtle for most of training. Due to the large number of inputs, each individual weight is relatively small, as only the total sum is relevant to the timing of the output spike. It should be noted that some weights continue to change throughout the whole of training. These weights, in particular, do not contribute significantly to any symbol in the training library and, therefore, have limited data when making weight adjustments.

      After about 10 epochs, the trained neural network will produce a spike within a 10 ps widow of the target time when the symbol is recognized, as shown in Fig. 3.

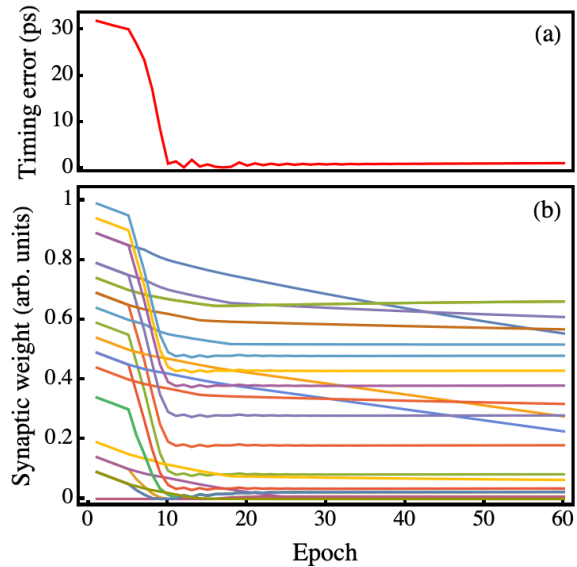

Figure 4. Simulations depicting the training process of an AFM SPAN. (a) Time difference between the target and actual spike times of the AFM SPAN's output over 60 epochs of training. (b) Each colored line illustrates the evolution of an individual synaptic weight from an input neuron to the AFM SPAN over 60 epochs of training.

      Several examples of incorrect symbols serving as input and the resulting output spikes are shown in Fig. 5. By spiking outside the target time window for any symbol other than the correct symbol, the neural network has high accuracy in recognizing the chosen symbol. Whether additional or missing pixels serve as the difference from the correct image does not matter in making a spike outside of the target time window.

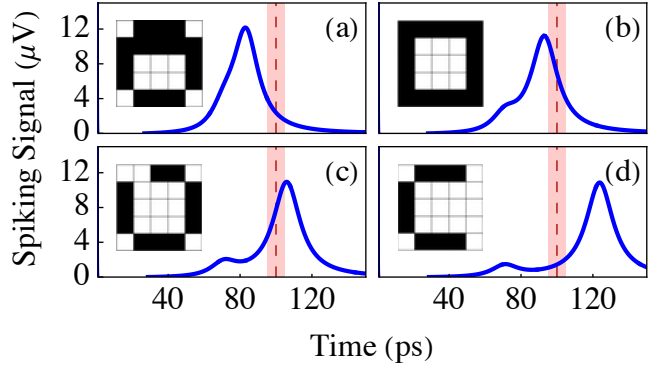

Figure 5. Simulation result of the output spikes of a trained AFM SPAN for different incorrect symbols, shown by insets. The dashed line represents the target time, and the red shading illustrates a 10 ps time window surrounding the target time. The simulated response for each incorrect symbol is outside the target time window, indicating that the inputted symbol is recognized to be not the correct symbol. (a, b) Input symbols have additional pixels. (c, d) Input symbols have missing pixels.

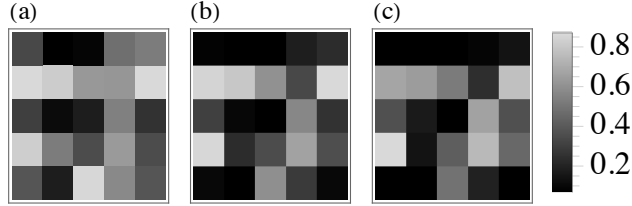

Figure 6. Distribution of weights connecting inputs to SPAN (a) at the beginning of training, (b) in the middle of training, (c) at the end of training.

To gain a more complete understanding of how weights change when training with the "Z" symbol, a distribution of weights is plotted in Fig. 6. Figure 6(a) shows the random distribution of weight at the beginning of training, and Fig. 6(c) shows the weights at the end of training, after 60 epochs. Figure 6(b), in contrast, shows the training in the middle of training after 10 epochs. It is evident from this sequence that as training progresses, the weights are adjusted in such a way that the "Z" symbol is reflected in the weight distribution. As there is little difference between Fig. 6(b) and Fig. 6(c), it can be assumed that the most significant training happens in the first 10 epochs.

Multiple AFM SPANs, trained to recognize different symbols, can be connected to the same layer of input neurons. This way, the SPAN trained to the input symbol will produce a spike within the target time window, while the others would spike outside it. With multiple SPANs all spiking at different times, the output can be unclear. Therefore, it would be convenient to clear the output by suppressing output spikes outside the target time window. This can be done by creating an additional output layer that consists of fixed synapses. The architecture of the neural network capable of suppressing unwanted outputs is shown in Fig. 7.

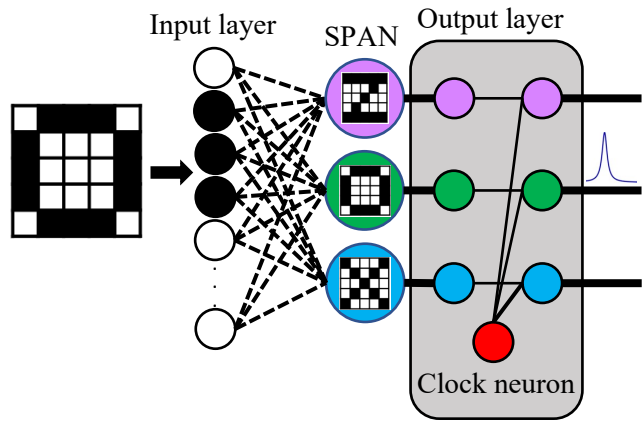

Figure 7. Architecture of a 3 AFM SPAN neural network. The input symbol is encoded in the spiking pattern of the input layer neurons. During training, the synaptic weights between the input neurons and each SPAN are adjusted individually as each SPAN is trained to recognize a different symbol. The weights of the output layer are weakened such that a single spike cannot propagate to the next neuron. The red neuron represents a clock neuron that spikes at the target time, combining with the SPAN corresponding to the recognized symbol to generate a sufficiently strong signal for the post-synaptic neuron to fire.

The input neurons are connected to three SPANs via trainable weights. These SPANs are each trained to recognize different symbols. The SPANs then serve as input to the output layer. The output layer's synapses have weak coupling, such that a single pre-synaptic spike is insufficient to cause a spike in the post-synaptic neuron. Two spikes must happen simultaneously to produce a strong enough signal for a post-synaptic spike.

The red neuron, shown in Fig. 7, is a clock neuron spiking at the target time. The clock neuron receives an input independent of the input layer's symbol. This independent input causes the clock neuron to generate a spike at the target time. Therefore, when a symbol is recognized, the SPAN will spike along with the clock neuron at the target time. These two signals are enough to overcome the weak coupling and cause the post-synaptic neuron to fire. The spikes from the SPANs that do not correspond to the input symbol would spike away from the target time, thus not combining with the clock neuron to cause an output spike. This output layer ensures that only the spike from the SPAN corresponding to the correct symbol is outputted. The output spiking signals of this neural network are shown in Fig. 8.

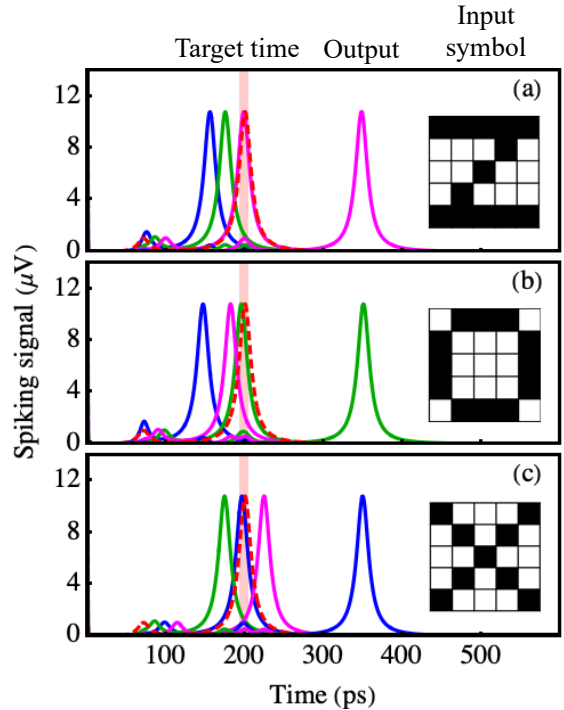

Figure 8. Simulated result of a 3 SPAN (pink, green, blue) neural network for 3 different input symbols (Z, O, X). The red shading represents a 10 ps time window surrounding the target time. The SPAN corresponding to the recognized symbol generates a spike within the target time window, along with the clock neuron (red dashed spike), resulting in a single output spike. The color of the output spike corresponds to the SPAN that was trained to recognize the inputted symbol.

The blue, green, and magenta spikes correspond to three different SPANs trained to three different symbols, while the red spike is the clock neuron spiking at the target time. At this time, there are spikes of the clock neuron and a single SPAN corresponding to the inputted symbol. These two spikes combine to send a single spike to the output, indicating which symbol has been recognized. Therefore, this output layer creates an output that clearly identifies the recognized symbol.

The simulated AFM neural networks are capable of recognizing symbols by producing a spike within a target time window (10 ps). The training time of the AFM networks for such relatively small images is very short, about 10 epochs with a 20-symbol library. Due to the high speed of AFM neurons (200 ps between inputting a symbol and the output neuron firing), this training session may last for only ~ 40 ns of real-time.

The energy consumption of a single AFM neuron, with dimensions described in Ref. [22] was calculated to be about $10^{-3}$ pJ per synaptic operation. This is a relatively low power consumption in comparison with other artificial spiking neurons. For example, a spin torque nano-oscillator-based neuron reports an energy consumption of 4.96 pJ per synaptic operation [30], and an all-CMOS neuron was reported to have a power consumption of 247 pJ per synaptic operation [31]. The low power consumption of AFM neurons is the result of the simple design of the AFM oscillator and the ultra-fast spikes that result from using AFM materials. To successfully train a single SPAN, the total energy consumption of the AFM neural network is 31.2 pJ.

**Conclusions**

Ultra-fast spiking artificial neurons built from AFM oscillators have a number of unique properties that can be harnessed to create simple neural networks with fixed synapses. However, in order to use AFM neurons in neuromorphic computing, a trainable neural network with variable synapses is required. The physical implementation of variable synapses is much more complex than fixed ones; therefore, in this work, we limit the number of trainable weights to reduce the complexity of the neural network. Namely, we numerically investigated the performance of the SPAN algorithm for the recognition of symbols encoded in a $5 \times 5$ binary grid.

Multiple SPANs, trained to different symbols, can be connected to the same inputs, thus providing multi-symbol recognition capabilities. With the addition of a fixed output layer suppressing spikes outside the target time window, the neural network will produce a single spike corresponding to the recognized symbol in just a few hundred picoseconds.

The use of the SPAN algorithm leads to a very simple, one-layer neural network. This neural network is limited in its ability to perform more complex tasks as a result of such simplifications. For example, the MNIST data set of handwritten digits, commonly used for neural network training, is encoded in $28 \times 28$ pixel grids. It is unlikely that a neural network such as the one studied here would be able to cope with inputs on this scale. The simple neural network used in this work demonstrates for the first time that AFM neurons are capable of being used for neuromorphic tasks such as pattern recognition. In order to advance the use of AFM neurons in neuromorphic computing, a more complex neural network and learning algorithm are required. However, even the simple network described in this work may find practical applications when high training, operation speed, and/or low consumed power are required.

**Acknowledgments**

This work was partially supported by the Air Force Office of Scientific Research (AFOSR) Multidisciplinary Research Program of the University Research Initiative (MURI), under Grant No. FA9550-19-1-0307.

**Data Availability**

Data may be made available at the request of the reader by contacting the authors at hbradley@oakland.edu.

**References**

[1] W. B. Levy and V. G. Calvert, "Communication consumes 35 times more energy than computation in the human cortex, but both costs are needed to predict synapse number," *Proc. Natl. Acad. Sci. U. S. A.*, vol. 118, no. 18, p. e2008173118, May 2021, doi: 10.1073/pnas.2008173118.

[2] "Big data needs a hardware revolution," *Nature*, vol. 554, no. 7691, pp. 145–146, Feb. 2018, doi: 10.1038/d41586-018-01683-1.

[3] J. Zhu, T. Zhang, Y. Yang, and R. Huang, "A comprehensive review on emerging artificial neuromorphic devices," *Appl. Phys. Rev.*, vol. 7, no. 1, p. 011312, Feb. 2020, doi: 10.1063/1.5118217.

[4] M. Davies *et al.*, "Loihi: A Neuromorphic Manycore Processor with On-Chip Learning," *IEEE Micro*, vol. PP, pp. 1–1, Jan. 2018, doi: 10.1109/MM.2018.112130359.

[5] J. Grollier, D. Querlioz, K. Y. Camsari, K. Everschor-Sitte, S. Fukami, and M. D. Stiles, "Neuromorphic spintronics," *Nat. Electron.*, vol. 3, no. 7, Art. no. 7, Jul. 2020, doi: 10.1038/s41928-019-0360-9.

[6] G. J. Lim, C. C. I. Ang, and W. S. Lew, "Spintronics for Neuromorphic Engineering," in *Emerging Non-volatile Memory Technologies: Physics, Engineering, and Applications*, W. S. Lew, G. J. Lim, and P. A. Dananjaya, Eds., Singapore: Springer, 2021, pp. 297–315. doi: 10.1007/978-981-15-6912-8_9.

[7] C. H. Marrows, J. Barker, T. A. Moore, and T. Moorsom, "Neuromorphic computing with spintronics," *Npj Spintron.*, vol. 2, no. 1, pp. 1–7, Apr. 2024, doi: 10.1038/s44306-024-00019-2.

[8] G. Finocchio *et al.*, "Roadmap for Unconventional Computing with Nanotechnology," *Nano Futur.*, 2024, doi: 10.1088/2399-1984/ad299a.

[9] J. Grollier, D. Querlioz, and M. D. Stiles, "Spintronic Nanodevices for Bioinspired Computing," *Proc. IEEE Inst. Electr. Electron. Eng.*, vol. 104, no. 10, p. 2024, Oct. 2016, doi: 10.1109/JPROC.2016.2597152.

[10] N. Hassan *et al.*, "Magnetic domain wall neuron with lateral inhibition," *J. Appl. Phys.*, vol. 124, no. 15, p. 152127, 2018.

[11] W. H. Brigner *et al.*, "Shape-Based Magnetic Domain Wall Drift for an Artificial Spintronic Leaky Integrate-and-Fire Neuron," *IEEE Trans. Electron Devices*, vol. 66, no. 11, pp. 4970–4975, Nov. 2019, doi: 10.1109/TED.2019.2938952.

[12] D. Wang *et al.*, "Spintronic leaky-integrate-fire spiking neurons with self-reset and winner-takes-all for neuromorphic computing," *Nat. Commun.*, vol. 14, no. 1, Art. no. 1, Feb. 2023, doi: 10.1038/s41467-023-36728-1.

[13] "A tunable magnetic skyrmion neuron cluster for energy efficient artificial neural network | IEEE Conference Publication | IEEE Xplore." Accessed: Apr. 05, 2023. [Online]. Available: https://ieeexplore.ieee.org/abstract/document/7927015/

[14] X. Chen *et al.*, "A compact skyrmionic leaky–integrate–fire spiking neuron device," *Nanoscale*, vol. 10, no. 13, pp. 6139–6146, Mar. 2018, doi: 10.1039/C7NR09722K.

[15] S. Li, W. Kang, Y. Huang, X. Zhang, Y. Zhou, and W. ZHAO, "Magnetic skyrmion-based artificial neuron device," *Nanotechnology*, vol. 28, p. 31LT01, Jul. 2017, doi: 10.1088/1361-6528/aa7af5.

[16] J. Torrejon *et al.*, "Neuromorphic computing with nanoscale spintronic oscillators," *Nature*, vol. 547, no. 7664, Art. no. 7664, Jul. 2017, doi: 10.1038/nature23011.

[17] A. Sengupta, P. Panda, P. Wijesinghe, Y. Kim, and K. Roy, "Magnetic Tunnel Junction Mimics Stochastic Cortical Spiking Neurons," *Sci. Rep.*, vol. 6, no. 1, Art. no. 1, Jul. 2016, doi: 10.1038/srep30039.

[18] M. Zahedinejad et al., "Two-dimensional mutually synchronized spin Hall nano-oscillator arrays for neuromorphic computing," *Nat. Nanotechnol.*, vol. 15, no. 1, Art. no. 1, Jan. 2020, doi: 10.1038/s41565-019-0593-9.

[19] J. Cai et al., "Voltage-Controlled Spintronic Stochastic Neuron Based on a Magnetic Tunnel Junction," *Phys. Rev. Appl.*, vol. 11, no. 3, p. 034015, Mar. 2019, doi: 10.1103/PhysRevApplied.11.034015.

[20] D. R. Rodrigues et al., "Spintronic Hodgkin-Huxley-Analogue Neuron Implemented with a Single Magnetic Tunnel Junction," *Phys. Rev. Appl.*, vol. 19, no. 6, p. 064010, Jun. 2023, doi: 10.1103/PhysRevApplied.19.064010.

[21] R. Khymyn et al., "Ultra-fast artificial neuron: generation of picosecond-duration spikes in a current-driven antiferromagnetic auto-oscillator," *Sci. Rep.*, vol. 8, no. 1, Art. no. 1, Oct. 2018, doi: 10.1038/s41598-018-33697-0.

[22] H. Bradley et al., "Artificial neurons based on antiferromagnetic auto-oscillators as a platform for neuromorphic computing," *AIP Adv.*, vol. 13, no. 1, p. 015206, Jan. 2023, doi: 10.1063/5.0128530.

[23] G. Tanaka et al., "Recent advances in physical reservoir computing: A review," *Neural Netw.*, vol. 115, pp. 100–123, Jul. 2019, doi: 10.1016/j.neunet.2019.03.005.

[24] A. Mohemmed, S. Schliebs, S. Matsuda, and N. Kasabov, "Training spiking neural networks to associate spatio-temporal input–output spike patterns," *Neurocomputing*, vol. 107, pp. 3–10, May 2013, doi: 10.1016/j.neucom.2012.08.034.

[25] A. Mohemmed, S. Schliebs, S. Matsuda, and N. Kasabov, "Method for Training a Spiking Neuron to Associate Input-Output Spike Trains," in *Engineering Applications of Neural Networks*, L. Iliadis and C. Jayne, Eds., in IFIP Advances in Information and Communication Technology. Berlin, Heidelberg: Springer, 2011, pp. 219–228. doi: 10.1007/978-3-642-23957-1_25.

[26] G. Tanaka et al., "Recent advances in physical reservoir computing: A review," *Neural Netw.*, vol. 115, pp. 100–123, Jul. 2019, doi: 10.1016/j.neunet.2019.03.005.

[27] R. Khymyn, I. Lisenkov, V. Tiberkevich, B. A. Ivanov, and A. Slavin, "Antiferromagnetic THz-frequency Josephson-like Oscillator Driven by Spin Current," *Sci. Rep.*, vol. 7, no. 1, Art. no. 1, Mar. 2017, doi: 10.1038/srep43705.

[28] K. P. McKenna and G. J. Morgan, "Quantum simulations of spin-relaxation and transport in copper," *Eur. Phys. J. B*, vol. 59, no. 4, pp. 451–456, Oct. 2007, doi: https://doi.org/10.1140/epjb/e2007-00305-2.

[29] A. Mohemmed, S. Schliebs, S. Matsuda, and N. Kasabov, "Training spiking neural networks to associate spatio-temporal input–output spike patterns," *Neurocomputing*, vol. 107, pp. 3–10, May 2013, doi: 10.1016/j.neucom.2012.08.034.

[30] K. Yogendra, D. Fan, and K. Roy, "Coupled Spin Torque Nano Oscillators for Low Power Neural Computation," *IEEE Trans. Magn.*, vol. 51, no. 10, pp. 1–9, Oct. 2015, doi: 10.1109/TMAG.2015.2443042.

[31] P. Livi and G. Indiveri, "A current-mode conductance-based silicon neuron for address-event neuromorphic systems," in *2009 IEEE International Symposium on Circuits and Systems*, May 2009, pp. 2898–2901. doi: 10.1109/ISCAS.2009.5118408.

**Author contributions statement**

HB is responsible for numerical simulations, data analysis and writing the manuscript. VT contributed to formulating the program and provided overall supervision. All authors participated in reviewing and editing the manuscript.